\documentclass{article}

\usepackage[nonatbib,final]{nips_2017}
\usepackage[numbers,sort]{natbib}
\usepackage{amsmath, amssymb}
\usepackage{MarkMathCmds}
\newcommand{\cov}{\textrm{cov}}

\newcommand{\Kuu}{\mathbf{K}_{\mathbf{uu}}}
\newcommand{\Kuf}{\mathbf{K}_{\mathbf {uf}}}
\newcommand{\Kfu}{\mathbf{K}_{\mathbf {fu}}}
\newcommand{\Kff}{\mathbf{K}_{\mathbf {ff}}}

\newcommand{\bk}{\mathbf k}
\newcommand{\bm}{\mathbf m}

\newcommand{\by}{\mathbf y}
\newcommand{\bx}{\mathbf x}
\newcommand{\bu}{\mathbf u}

\newcommand{\bS}{\mathbf S}
\newcommand{\bZ}{\mathbf Z}
\newcommand{\bz}{\mathbf z}

\newcommand{\dee}{\,\textrm d}
\newcommand{\given}{\,|\,}

\newcommand{\cGP}{\mathcal{GP}}

\newcommand{\cN}{\mathcal{N}}

\usepackage[utf8]{inputenc} 
\usepackage[T1]{fontenc}    
\usepackage{hyperref}       
\usepackage{url}            
\usepackage{booktabs}       
\usepackage{amsfonts}       
\usepackage{nicefrac}       
\usepackage{microtype}      
\usepackage{tikz, pgfplots, pgfplotstable}
\usepgfplotslibrary{groupplots}
\pgfplotsset{grid style=none}
\usetikzlibrary{pgfplots.colormaps}
\newlength\figurewidth
\newlength\figureheight
\usepackage{subcaption}

\title{Convolutional Gaussian Processes}
\author{
  Mark van der Wilk \\
  Department of Engineering\\
  University of Cambridge, UK\\
  \texttt{mv310@cam.ac.uk} \\
  \And
  Carl Edward Rasmussen\\
  Department of Engineering\\
  University of Cambridge, UK\\
  \texttt{cer54@cam.ac.uk} \\
  \And
  James Hensman\\
  prowler.io\\
  Cambridge, UK\\
  \texttt{james@prowler.io} \\
}
\begin{document} 
\maketitle

\begin{abstract} 
We present a practical way of introducing convolutional structure into Gaussian processes, making them more suited to high-dimensional inputs like images. The main contribution of our work is the construction of an inter-domain inducing point approximation that is well-tailored to the convolutional kernel. This allows us to gain the generalisation benefit of a convolutional kernel, together with fast but accurate posterior inference. We investigate several variations of the convolutional kernel, and apply it to MNIST and CIFAR-10, which have both been known to be challenging for Gaussian processes. We also show how the marginal likelihood can be used to find an optimal weighting between convolutional and RBF kernels to further improve performance. We hope that this illustration of the usefulness of a marginal likelihood will help automate discovering architectures in larger models.
\end{abstract} 

\section{Introduction} 
\label{par:intro}
Gaussian process (GPs) \citep{williams2006gaussian} can be used as a flexible prior over functions, which makes them an elegant building block in Bayesian nonparametric models. In recent work, there has been much progress in addressing the computational issues preventing GPs from scaling to large problems \citep{titsias2009variational, hensman2013gaussian, matthews2015sparse, hensman2015mcmc}. However, orthogonal to being able to algorithmically handle large quantities of data is the question of how to build GP models that generalise well. The properties of a GP prior, and hence its ability to generalise in a specific problem, are fully encoded by its covariance function (or kernel). Most common kernel functions rely on rather rudimentary and local metrics for generalisation, like the euclidean distance. This has been widely criticised, notably by \citet{bengio2009deep}, who argued that deep architectures allow for more non-local generalisation. While deep architectures have seen enormous success in recent years, it is an interesting research question to investigate what kind of non-local generalisation structures \emph{can} be encoded in shallow structures like kernels, while preserving the elegant properties of GPs. 

Convolutional structures have non-local influence and have successfully been applied in neural nets to improve generalisation for image data \citep[see e.g.][]{LeNet1998, AlexNet2012}. In this work, we investigate how Gaussian processes can be equipped with convolutional structures, together with accurate approximations that make them applicable in practice. A previous approach by  \citet{wilson2016stochastic} transforms the inputs to a kernel using a convolutional neural network. This is valid since applying a deterministic transformation to kernel inputs results in a valid kernel \citep[see e.g.][]{williams2006gaussian, calandra2016manifold}, with the (many) parameters of the transformation becoming kernel hyperparameters. We stress that our approach is different in that the process itself is convolved, which does not require the introduction of additional parameters. Although our method does have inducing points that play a similar role to the filters in a convnet, these are variational parameters and thus protected from over-fitting.

\section{Background} 
\label{par:background}
Interest in Gaussian processes in the machine learning community started with the realisation that a shallow but infinitely wide network with Gaussian weights was a Gaussian process \citep{neal1996bayesian} -- a nonparametric model with analytically tractable posteriors and marginal likelihoods. This gives two main desirable properties. Firstly, the posterior gives error bars, which, combined with having an infinite number of basis functions, results in sensibly large error bars far from the data (see \citet[fig. 5]{quinonero2005unifying} for a useful illustration). Secondly, the marginal likelihood can be used to select kernel hyperparameters. The main drawback is an $\BigO\left(N^3\right)$ computational cost for $N$ observations. Because of this, much attention over recent years has been devoted to scaling GP inference to large datasets through sparse approximations \citep{Seeger2003dtc, snelson2005fitc, titsias2009variational}, minibatch-based optimisation \citep{hensman2013gaussian}, exploiting structure in the covariance matrix \citep[e.g.][]{wilson2015kernel} and Fourier methods \citep{lazaro2010sparse, hensman2016variational}.

In this work, we adopt the variational framework for approximation in GP models, because it can simultaneously handle computational speed-up to $\BigO\left(NM^2\right)$ (with $M << N$) through sparse approximations \citep{titsias2009variational} and approximate posteriors due to non-Gaussian likelihoods \citep{opper2009variational}. The variational choice is both elegant and practical: it can be shown that the variational objective minimises the KL divergence across the entire latent process \citep{matthews2015sparse, matthews2016phd}, which guarantees that the exact model will be approximated given enough resources. Other methods, such as EP/FITC \citep{snelson2005fitc,hernandez2016scalable,Bui2016unifying,villacampa2017gp-ep-multi}, can be seen as approximate models that do not share this property, leading to behaviour that would not be expected from the model that is to be approximated \citep{bauer2016understanding}. It is worth noting however, that our method for convolutional GPs is general to the objective function used to train the model, and can plug in without modification to EP style objective functions.

\subsection{Gaussian variational approximation}
We adopt the popular choice of combining a sparse GP approximation with a Gaussian assumption, using a variational objective as introduced in \citep{hensman2015scalable}. The model is written
\begin{align}
  f(\cdot)\given \theta &\sim \cGP\left(0, \, k(\cdot,\cdot)\right)\,, \label{eq:prior_GP} \\
  y_i\given f, \vx_i &\stackrel{iid}{\sim} p(y_i\given f(\bx_i))\,,\label{eq:likelihood}
\end{align}
where $p(y_i\given f(\bx_i))$ is some non-Gaussian likelihood, for example a Bernoulli distribution through a probit link function for classification. The kernel parameters $\theta$ are to be estimated by approximate maximum likelihood, we drop them from the notation hereon. We choose the approximate posterior as \citet{titsias2009variational} to be a GP with its marginal distribution specified at $M$ ``inducing inputs'' $\bZ = \{\bz_m\}_{m=1}^M$. Denoting the value of the GP at those points as $\bu = \{f(\bz_m)\}_{m=1}^M$, the approximate posterior process is constructed from the specified marginal, and the prior conditional\footnote{The construction of the approximate posterior can alternatively be seen as a GP posterior to a regression problem, where the $q(\vu)$ indirectly specifies the likelihood. Variational inference will then adjust the inputs and likelihood of this regression problem to make the approximation close to the true posterior in KL divergence.}
\begin{align}
	\bu &\sim \cN\big(\bm, \, \bS\big)\,,\label{eq:q_of_u}\\
	f(\cdot)\given \bu &\sim \cGP\big(\bk_\bu(\cdot)^\top\Kuu^{-1}\bu, \, k(\cdot,\cdot) - \bk_\bu(\cdot)^\top\Kuu^{-1}\bk_\bu(\cdot)\big)\,.\label{eq:q_f_given_u}
\end{align}
The vector-valued function $\bk_\bu(\cdot)$ gives the covariance between $\bu$ and the remainder of $f$, and is constructed from the kernel: $\bk_\bu(\cdot)=\big[k(\bz_m, \cdot)\big]_{m=1}^M$. The matrix $\Kuu$ is the prior covariance of $\bu$.
The variational parameters $\bm$, $\bS$ and $\bZ$ are then optimised with respect to the ELBO:
\begin{equation}
	\textsc{ELBO} = \sum_i \Exp{q(f(\vx_i))}{\log p(y_i\given f(\bx_i))} - \textsc{KL}[q(\bu)||p(\bu)]\,.
\end{equation}
Here, $q(\bu)$ is the density of $\bu$ associated with equation \eqref{eq:q_of_u}, and $p(\bu)$ is the prior density from \eqref{eq:prior_GP}. Expectations are taken with respect to the marginals of the posterior approximation, given by
\begin{align}
    q(f(\bx_i)) &= \cN\left(\mu_i, \sigma^2_i\right)\,,\label{eq:q_f} \\
	\mu_i &= \bk_\bu(\bx_i)^\top\Kuu^{-1}\bm\,,\label{eq:q_mean} \\
    \sigma^2_i &= k(\bx_i, \bx_i)+ \Kfu\Kuu^{-1}(\bS-\Kuu)\Kuu^{-1}\Kuf\,.\label{eq:q_var}
\end{align}
The matrices $\Kuu$ and $\Kfu$ are obtained by evaluating the kernel as $k(\vz_m, \vz_{m'})$ and $k(\vx_n, \vz_m)$ respectively. The Kullback-Leibler term of the ELBO is tractable, whilst the expectation term can be computed using one-dimensional quadrature. The form of the ELBO means that stochastic optimization using minibatches is applicable. A full discussion of the methodology is given by \citet{matthews2016phd}. We optimise the ELBO instead of the marginal likelihood to find the hyperparameters.

\subsection{Inter-domain variational GPs}
Inter-domain Gaussian processes \citep{figueiras2009inter} work by replacing the variables $\bu$, which we have above assumed to be observations of the function at the inducing inputs $\bZ$, with more complicated variables made by some linear operator on the function. Using linear operators ensures that the inducing variables $\vu$ are still jointly Gaussian with the other points on the GP. Implementing inter-domain inducing variables can therefore be a drop-in replacement to inducing points, requiring only that the appropriate (cross-) covariances $\Kfu$ and $\Kuu$ are used.

The key advantage of the inter domain approach is that the effective basis functions of the sparse approximation can be made more flexible. The effective basis functions of the approximate posterior mean \eqref{eq:q_mean} are given by $\bk_\bu(\cdot)$. In an inter-domain approach, these basis functions can be constructed to take an alternative form by manipulating the linear operator which constructs $\bu$. For example, \citet{hensman2016variational} used a Fourier transform to construct $\bu$ variables in the Fourier domain.

Inter-domain inducing variables are usually constructed using a weighted integral of the GP:
\begin{equation}
    u_m = \int \phi(\bx, \bz_m) f(\bx) \dee \bx\,,
\end{equation}
where the weighting function depends on some parameters $\bz_m$. The covariance between the inducing variable $u_m$ and a point on the function is then
\begin{equation}
    \cov(u_m, f(\bx_n)) = k(\vz_m, \vx_n) =  \int \phi(\bx, \bz_m) k(\bx, \bx_n) \dee \bx\,,
\end{equation}
and the covariance between two inducing variables is
\begin{equation}
    \cov(u_m, u_{m'}) = k(\vz_m, \vz_{m'}) = \int\int \phi(\bx, \bz_m)\phi(\bx', \bz_{m'}) k(\bx, \bx') \dee \bx \dee \bx'\,.
\end{equation}

Using inter-domain inducing variables in the variational framework is straightforward if the above integrals are tractable. The results are substituted for the kernel entries in equations \eqref{eq:q_mean} and \eqref{eq:q_var}. 


Our proposed method will be an inter-domain approximation in the sense that the inducing input space is different from the input space of the kernel. However, instead of relying on an integral transformation of the GP, we construct the inducing variables $\vu$ alongside the new kernel such that the effective basis functions contain a convolution operation.


\subsection{Additive GPs}
We would like to draw attention to previously studied additive models \citep{durrande2011additive, duvenaud2011additive}, in order to highlight the similarity with the convolutional kernels we will introduce later. Additive models construct a prior GP as a sum of functions over subsets of the input dimensions, resulting in a kernel with the same additive structure. For example, summing over each input dimension, we get
\begin{align}
	f(\vx) = \sum_i f_i(\vx[i]) \implies k(\bx, \bx') = \sum_i k_0(\bx[i], \bx'[i])\,. \label{eq:additive-kern}
\end{align}
This kernel exhibits some non-local generalisation, as the relative function values along one dimension will be the same regardless of the input along other dimensions. In practice, this specific additive model is rather too restrictive to fit data well, since it assumes that all variables affect the response $\by$ independently. At the other extreme, the popular squared exponential kernel allows interactions between all dimensions, but this turns out to be not restrictive enough: for high dimensional problems we need to impose {\em some} restriction on the form of the function.

In this work, we build an additive kernel inspired by the convolution operator found in convnets. The same function is applied to patches from the input, which allows adjacent pixels to interact, but imposes an additive structure otherwise.



\section{Convolutional Gaussian Processes} 
\label{par:cgp}
In the next few sections, we will introduce several variants of the Convolutional Gaussian process, and illustrate its properties using toy and real datasets. Our main contribution is showing that convolutional structure can be embedded in kernels, and that they can be used within the framework of nonparametric Gaussian process approximations. We do so by constructing the kernel in tandem with a suitable domain to place the inducing variables in. Implementation\footnote{Ours can be found on \url{https://github.com/markvdw/convgp}, together with code for replicating the experiments, and trained models. It is based on GPflow \citep{gpflow}, allowing utilisation of GPUs.} requires minimal changes to existing implementations of sparse variational GP inference, and can leverage GPU implementations of convolution operations (see appendix). In the appendix we also describe how the same inference method can be applied to kernels with general invariances.


\paragraph{Convolutional kernel construction} We construct a convolutional GP by starting with a \emph{patch-response function}, $g: \Reals^E \to \Reals$, mapping from patches of size $E$. For images of size $D = W\times H$, and patches of size $E = w\times h$, we get a total of $P = (W - w + 1) \times (H - h + 1)$ patches. We can start by simply making the overall function from the image $f: \Reals^D \to \Reals$ the sum of all patch responses. If $g(\cdot)$ is given a GP prior, a GP prior will also be induced on $f(\cdot)$:
\begin{align}
g &\sim \GP\left(0, k_g(\vz, \vz')\right)\,,\quad
f(\vx) = \sum_p g(\vx^{[p]}) \,,\\
    \implies f &\sim \GP\left(0, \sum_{p=1}^P\sum_{p'=1}^P k_g\left(\vx^{[p]}, \vx'^{[p']}\right)\right)\,, \label{eq:convkern}
\end{align}
where $\vx^{[p]}$ indicates the $p^\textrm{th}$ patch of the vector $\vx$. This construction is reminiscent of the additive models discussed earlier, since a function is applied to subsets of the input. However, in this case, the \emph{same} function $g(\cdot)$ is applied to all input subsets. This allows distant patches to inform the value of the patch-response function.

\paragraph{Comparison to convnets} This approach is similar in spirit to convnets. Both methods start with a function that is applied to each patch. In the construction above, we introduce a single patch-response function $g(\cdot)$ that is a non-linear and nonparametric. Convnets, on the other hand, rely on many linear filters, followed by a non-linearity. The flexibility of a single convolutional layer is controlled by the number of filters, while depth is important in order to allow for enough non-linearity. In our case, adding more non-linear filters to the construction of $f(\cdot)$ does not increase capacity to learn. The patch responses of the multiple filters would be summed, resulting in simply a summed kernel for the prior over $g$.

\paragraph{Computational issues} Similar kernels have been proposed in various forms \citep{mairal2014ckn, pandey2014stretching}, but have never been applied directly in GPs, probably due to the prohibitive costs. Direct implementation of a GP using $k_f$ would be infeasible not only due to the usual cubic cost w.r.t.~the number of data points, but also due to it requiring $P^2$ evaluations of $k_g$ \emph{per element of $\Kff$}. For MNIST with patches of size 5, $P^2 \approx 3.3 \cdot 10^5$, resulting in the kernel evaluations becoming a significant bottleneck. Sparse inducing point methods require $M^2 + NM$ kernel evaluations of $k_f$, which would still be infeasible with the large number of patches. Luckily, a bigger improvement is possible.

\subsection{Translation invariant convolutional GP}
Here we introduce the simplest version of our method. We start with the construction from section \ref{par:cgp}. In order to obtain a tractable method, we want to approximate the true posterior using a small set of inducing points. The main idea is to place these inducing points in the input space of \emph{patches}, rather than images. This corresponds to using inter-domain inducing points. In order to use this approximation we simply need to find the appropriate inter-domain (cross-) covariances $\Kuu$ and $\Kfu$, which are easily found from the construction of the convolutional kernel in equation \ref{eq:convkern}:
\begin{gather}
k_{fu}(\vx, \vz) = \Exp{g}{f(\vx) g(\vz)} = \Exp{g}{\sum_p g(\vx^{[p]}) g(\vz)} = \sum_p k_g\left(\vx^{[p]}, \vz\right) \label{eq:kfu} \,,
\end{gather}
\begin{gather}
k_{uu}(\vz, \vz') = \Exp{g}{g(\vz) g(\vz')} = k_g(\vz, \vz') \,.
\end{gather}
This improves on the computation from the standard inducing point method, since only covariances between the image patches and inducing patches are needed, allowing $\Kfu$ to be calculated with $NMP$ instead of $NMP^2$ kernel evaluations. Since $\Kuu$ now only requires the covariances between \emph{inducing patches}, its cost is $M^2$ instead of $M^2P^2$ evaluations. However, evaluating $\diag \left[\Kff\right]$ does still require $NP^2$ evaluations, although $N$ can be small when using minibatch optimisation. This brings the cost for computing the kernel matrices down significantly compared to the $\BigO\left(NM^2\right)$ cost of the calculation of the ELBO.

In order to highlight the capabilities of the new kernel, we now consider two toy tasks: classifying rectangles and distinguishing zeros from ones in MNIST.

\paragraph{Toy demo: rectangles}
The rectangles dataset is an artificial dataset containing $1200$ images of size $28\times 28$. Each image contains the outline of a randomly generated rectangle, and is labelled according to whether the rectangle has larger width or length. Despite its simplicity, the dataset is tricky for standard kernel-based methods, including Gaussian processes, because of the high dimensionality of the input, and the strong dependence of the label on multiple pixel locations.

\begin{figure}
\begin{subfigure}[t]{0.49\textwidth}
  \centering
  \includegraphics[width=\textwidth]{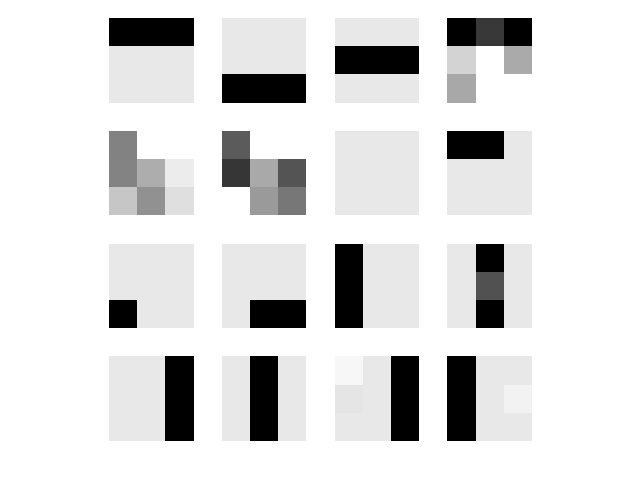}
  \vspace{-0.8cm}
  \caption{\label{fig:rectangles-patches}Rectangles dataset.}
  \end{subfigure}
\hfill
\begin{subfigure}[t]{0.49\textwidth}
  \centering
  \includegraphics[width=\textwidth]{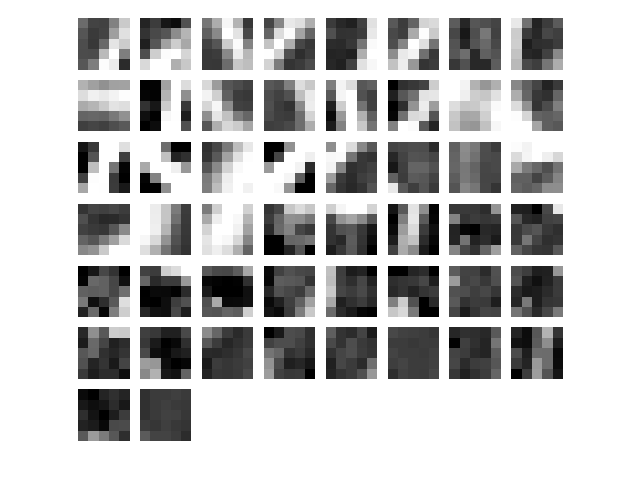}
  \vspace{-0.8cm}
  \caption{\label{fig:mnist01-patches}MNIST 0-vs-1 dataset.}
  \end{subfigure}
  \caption{The optimised inducing patches for the (approximately) translation invariant kernel. The inducing patches are sorted by the value of their corresponding inducing output, illustrating the evidence each patch has in favour of a class.}
\end{figure}

To tackle the rectangles dataset with the convolutional GP, we used a patch-size of $3\times 3$ and 16 inducing points initialised with uniform random noise. We optimised using Adam \citep{kingma2014adam} ($0.01$ learning rate \& 100 data points per minibatch) and obtained $1.4\%$ error and a negative log predictive probability (nlpp) of $0.055$ on the test set. For comparison, an RBF kernel with 1200 optimally placed inducing points, optimised with BFGS, gave $5.0\%$ error and an nlpp of $0.258$. The model is both better in terms of performance, and more compact in terms of inducing points. The model works because it is able to recognise and count vertical and horizontal bars in the patches. The locations of the inducing points quickly recognise the horizontal and vertical lines in the images -- see Figure \ref{fig:rectangles-patches}.

\paragraph{Illustration: Zeros vs ones MNIST}
We perform a similar experiment for classifying MNIST 0 and 1 digits. This time, we initialise using patches from the training data and use 50 inducing features, shown in figure \ref{fig:mnist01-patches}. Features in the top left are in favour of classifying a zero, and tend to be diagonal or bent lines, while features for ones tend to be blank space or vertical lines. We get $0.3\%$ errors.

\paragraph{Full MNIST}
Next, we turn to the full multi-class MNIST dataset. Our setup follows \citet{hensman2015mcmc}, with 10 independent latent GPs using the same convolutional kernel, and constraining $q(\vu)$ to a Gaussian (see section \ref{par:background}). It seems that this translation invariant kernel is too restrictive for this task, since the error rate converges at around $2.1\%$, compared to $1.9\%$ for the RBF kernel.

\subsection{Weighted convolutional kernels}
\label{sec:weightedconv}
We saw in the previous section that although the translation invariant kernel excelled at the rectangles task, it under-performed compared to the RBF on MNIST. Full translation invariance is too strong a constraint, which makes intuitive sense for image classification, as the same feature in different locations of the image can imply different classes. This can be remedied without leaving the family of Gaussian processes by weighting the response from each patch. Denoting again the underlying patch-based GP as $g$, the image-based GP $f$ is given by
\begin{equation}
  f(\vx) = \sum_p w_p g(\vx^{[p]})\,. \\
\end{equation}
The weights $\{w_p\}_{p=1}^P$ adjust the relative importance of the response for each location in the image. Only $k_f$ and $k_{fu}$ differ from the invariant case, and can be found to be:
\begin{gather}
    k_{f}(\vx, \vx) = \sum_{pq} w_p w_q k_g(\vx^{[p]}, \vx_q)\,, \\
    k_{fu}(\vx, \vz) = \sum_p w_p k_g(\vx^{[p]}, \vz)\,.
\end{gather}
The patch weights $\vw \in \Reals^P$ are considered to be kernel hyperparameters -- we optimise them with respect the the ELBO in the same fashion as the underlying parameters of the kernel $k_g$. This introduces $P$ hyperparameters into the kernel -- slightly less than the number of input pixels, which is how many hyperparameters an RBF kernel with automatic relevance determination would have.


\paragraph{Toy demo: rectangles} The errors in the previous section were caused by rectangles along the edge of the image, which contained bars which only contribute once to the classification score. Bars in the centre contribute in multiple patches. The weighting allows some up-weighting of patches along the edge. This results in near-perfect classification, with no classification errors and an nlpp of $0.005$.

\paragraph{Full MNIST} The weighting causes a significant reduction in error over the translation invariant and RBF kernels (table \ref{tab:mnist} \& figure \ref{fig:mnist}). The weighted convolution kernel obtains $1.22\%$ error -- a significant improvement over $1.9\%$ for the RBF kernel \citep{hensman2015mcmc}. \citet{autogp} report $1.55\%$ error using an RBF kernel, but using a Leave-One-Out objective for finding the hyperparameters.



\subsection{Does convolution capture everything?}
\label{sec:convplusrbf}
As discussed earlier, the additive nature of the convolution kernel places constraints on the possible functions in the prior. While these constraints have been shown to be useful for classifying MNIST, the guarantee of the RBF of enough capacity to model well in the large data limit, is lost: convolutional kernels are not universal \citep{steinwart2001,sriperumbudur2011kernelrelations} in the image input space, despite being nonparametric. This places convolutional kernels in a middle ground between parametric and universal kernels (see the appendix for a discussion). 
A kernel that \emph{is} universal \emph{and} has some amount of convolutional structure can be obtained by summing an RBF component: $f(\vx) = f_{conv}(\vx) + f_{rbf}(\vx)$. This allows the universal RBF to model any residuals that the convolutional structure cannot explain. We use the marginal likelihood to automatically weigh how much of the process should be explained by each of the components, in the same way as is done in other additive models \citep{duvenaud2013structure, duvenaud2011additive}.

Inference in such a model is straightforward under the usual inducing point framework -- it requires only evaluating the sum of kernels. The case considered here is more complicated since we want the inducing inputs for the RBF to lie in the space of images, while we want to use inducing patches for the convolutional kernel. This forces us to use a slightly different form for the approximating GP, representing the inducing inputs and outputs separately, as
\begin{gather}
\begin{bmatrix}\vu_{conv} \\ \vu_{rbf}\end{bmatrix} \sim \NormDist{\begin{bmatrix}\vmu_{conv} \\ \vmu_{rbf}\end{bmatrix}, \bS} \,, \\
f(\cdot)\given \bu = f_{conv}(\cdot)\given \bu_{conv} + f_{rbf}(\cdot)\given \bu_{rbf} \,.
\end{gather}
The variational lower bound changes only through the equations \eqref{eq:q_mean} and \eqref{eq:q_var}, which now must contain contributions of the two component Gaussian processes. If covariances in the posterior between $f_{conv}$ and $f_{rbf}$ are to be allowed, $\bS$ must be a full rank $2M\times 2M$ matrix. A mean-field approximation can be chosen as well, in which case $\bS$ can be $M\times M$ block-diagonal, saving some parameters. Note that regardless of which approach is chosen, the largest matrix to be inverted is still $M\times M$, as $\vu_{conv}$ and $\vu_{rbf}$ are independent in the prior (see the appendix for more details).

\begin{figure}[htpb]
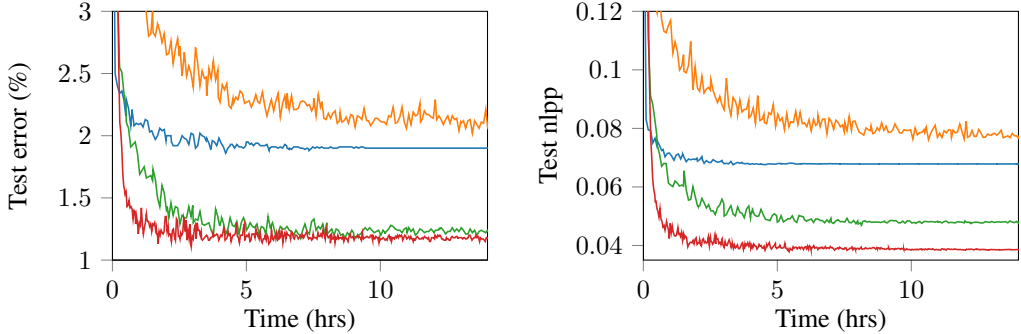

\setlength\figurewidth{0.47\textwidth}
\setlength\figureheight{0.35\columnwidth}
  \begin{subfigure}[t]{0.5\textwidth}
    \centering\input{figures/mnist-acc.tikz}
  \end{subfigure}
  \begin{subfigure}[t]{0.5\textwidth}
    \centering\input{figures/mnist-nlpp.tikz}
  \end{subfigure}
    \caption{\label{fig:mnist}Test error (left) and negative log predictive probability (nlpp, right) for MNIST, using RBF (blue), translation invariant convolutional (orange), weighted convolutional (green) and weighted convolutional + RBF (red) kernels.}
\end{figure}

\paragraph{Full MNIST} By adding an RBF component, we indeed get an extra reduction in error and nlpp from $1.22\%$ to $1.17\%$ and $0.048$ to $0.039$ respectively (table \ref{tab:mnist} \& figure \ref{fig:mnist}). The variances for the convolutional and RBF kernels are $14.3$ and $0.011$ respectively, showing that the convolutional kernel explains most of the variance in the data.

\begin{table}[h!]
\centering
\begin{tabular}{ c c c c }
Kernel & M & Error (\%) & NLPP \\
\hline
Invariant & 750 & $2.08\%$ & $0.077$ \\
RBF & 750 & $1.90\%$ & 0.068 \\
Weighted & 750 & $1.22\%$ & $0.048$ \\
Weighted + RBF & 750 & $1.17\%$ & 0.039 \\
\end{tabular}
\vspace{0.3cm}
\caption{Final results for MNIST.}
\label{tab:mnist}
\end{table}

\subsection{Convolutional kernels for colour images}
Our final variants of the convolutional kernel handle images with multiple colour channels. The addition of colour presents an interesting modelling challenge, as the input dimensionality increases significantly, with a large amount of redundant information. 
As a baseline, the weighted convolutional kernel from section \ref{sec:weightedconv} can be used by taking all patches from each colour channel together, resulting in $C$ times more patches. This kernel can only account for linear interactions between colour channels through the weights, and is also constrained to give the same patch response regardless of the colour channel. A step up in flexibility would be to define $g(\cdot)$ to take a $w\times h \times C$ patch with all $C$ colour channels. This trades off increasing the dimensionality of the patch-response function input with allowing it to learn non-linear interactions between the colour channels. We call this the \emph{colour-patch variant}. A middle ground that does not increase the dimensionality as much, is to use a different patch-response function $g_c(\cdot)$ for each colour channel. We will refer to this as the \emph{multi-channel} convolutional kernel. We construct the overall function $f$ as
\begin{align}
f(\vx) = \sum_{p=1}^P\sum_{c=1}^C w_{pc} g_c\left(\vx^{[pc]}\right) \,.
\end{align}
For this variant, inference becomes similar to section \ref{sec:convplusrbf}, although for a different reason. While all $g_c$s can use the same inducing patch inputs, we need access to each $g_c(\vx^{[pc]})$ separately in order to fully specify $f(\vx)$. This causes us to require separate inducing outputs for each $g_c$. In our approximation, we share the inducing inputs, while, as was done in section \ref{sec:convplusrbf}, representing the inducing outputs separately. The equations for $f(\cdot)|\vu$ are changed only through the matrices $\Kfu$ and $\Kuu$ being $N\times MC$ and $MC\times MC$ respectively. Given that the $g_c(\cdot)$ are independent in the prior, and the inducing inputs are constrained to be the same, $\Kuu$ is a block-diagonal repetition of $k_g\left(\vz_m, \vz_{m'}\right)$. All the elements of $\Kfu$ are given by
\begin{align}
k_{fg_c}(\vx, \vz) &= \Exp{\{g_c\}_{c=1}^C}{\sum_{p} w_{pc} g_c\left(\vx^{[pc]}\right) g_c(\vz)} = \sum_{p} w_{pc} k_g(\vx^{[pc]}, \vz) \,.
\end{align}
As in section \ref{sec:convplusrbf}, we have the choice to represent a full $CM\times CM$ covariance matrix for all inducing variables $\vu$, or go for a mean-field approximation requiring only $C$ $M\times M$ matrices. Again, both versions require no expensive matrix operations larger than $M\times M$ (see appendix).

Finally, a simplification can be made in order to avoid representing $C$ patch-response functions. If the weighting of each of the colour channels is constant w.r.t.~the patch location (i.e.~$w_{pc} = w_p w_c$), the model is equivalent to using a patch-response function with an additive kernel:
\begin{gather}
f(\vx) = \sum_p w_p \sum_c w_c g_c(\vx^{[pc]}) = \sum_p w_p \tilde{g}(\vx^{[pc]}) \label{eq:factorisedwpwc} \,, \\
\tilde{g}(\cdot) \sim \GP\left(0, \sum_c w_c k_c(\cdot, \cdot)\right) \,.
\end{gather}


\paragraph{CIFAR-10}
We conclude the experiments by an investigation of CIFAR-10 \citep{cifar10}, where $32\times 32$ sized RGB images are to be classified. We use a similar setup to the previous MNIST experiments, by using $5\times 5$ patches. Again, all latent functions share the same kernel for the prior, including the patch weights. We compare an RBF kernel to 4 variants of the convolutional kernel: the baseline ``weighted'', the colour-patch, the colour-patch variant with additive structure (equation \ref{eq:factorisedwpwc}), and the multi-channel with mean-field inference. All models use 1000 inducing inputs and are trained using Adam. Due to memory constraints on the GPU, a minibatch size of 40 had to be used for the weighted, additive and multi-channel models.

\begin{figure}[htpb]
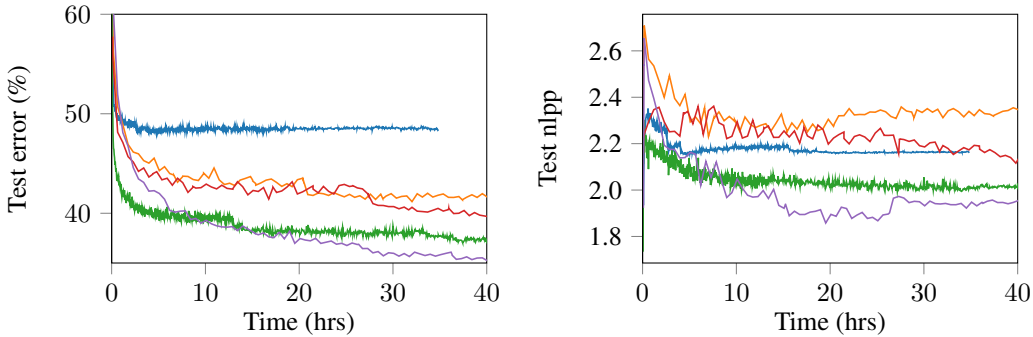

\setlength\figurewidth{0.47\textwidth}
\setlength\figureheight{0.35\columnwidth}
  \begin{subfigure}[t]{0.5\textwidth}
    \centering\input{figures/cifar-acc.tikz}
  \end{subfigure}
  \begin{subfigure}[t]{0.5\textwidth}
    \centering\input{figures/cifar-nlpp.tikz}
  \end{subfigure}
    \caption{\label{fig:cifar}Test error (left) and nlpp (right) for CIFAR-10, using RBF (blue), baseline weighted convolutional (orange), full-colour weighted convolutional (green), additive (red), and multi-channel (purple).}
\end{figure}

Test errors and nlpps during training are shown in figure \ref{fig:cifar}. Any convolutional structure significantly improves classification performance, with colour interactions seeming particularly important, as the best performing model is the multi-channel GP. The final error rate of the multi-channel kernel was $35.4\%$, compared to $48.6\%$ for the RBF kernel. While we acknowledge that this is far from state of the art using deep nets, it is a significant improvement over existing Gaussian process models, including the $44.95\%$ error reported by \citet{autogp}, where an RBF kernel was used together with their leave-one-out objective for the hyperparameters. This improvement is orthogonal to the use of a new kernel.


\vspace{-0.1cm}


\section{Conclusion} 
We introduced a method for efficiently using convolutional structure in Gaussian processes, akin to how it has been used in neural nets. Our main contribution is showing how placing the inducing inputs in the space of patches gives rise to a natural \emph{inter-domain} approximation that fits in sparse GP approximation frameworks. We discuss several variations of convolutional kernels and show how they can be used to push the performance of Gaussian process models on image datasets. Additionally, we show how the marginal likelihood can be used to assess to what extent a dataset can be explained with only convolutional structure. We show that convolutional structure is not sufficient, and that performance can be improved by adding a small amount of ``fully connected'' (RBF). The ability to do this, and automatically tune the hyperparameters is a real strength of Gaussian processes. It would be great if this ability could be incorporated in larger or deeper models as well.


\bibliographystyle{unsrtnat}
\setlength\bibsep{0pt}
{\small\bibliography{references}}

\end{document}


\maketitle

\appendix
\appendixpageoff

\section{Implementation}
\subsection{Inter-domain approximations}
Implementation of convolutional kernels requires only straightforward modifications of existing code that implements an inducing variable GP approximation. All inducing point methods like FITC \citep{snelson2005fitc}, variations on (Power-) EP \citep{bui2014pep} or variational free energy \citep{titsias2009variational, hensman2015mcmc} rely on the covariance of inducing variables and between observations and inducing variables:
\begin{align}
\Kuu &= \Cov\left[\vu\vu\transpose\right] \,, \\
\Kfu &= \Cov\left[\vf\vu\transpose\right] \,.
\end{align}
In normal inducing point approximations, $\vu$ and $\vf$ are simply evaluations of the latent GP of interest:
\begin{align}
[\vf]_n = f_n &= f(\vx_n) \,, && [\vu]_m = u_m = f(\vz_m) \,.
\end{align}
The resulting covariances are simply evaluations of the kernel of the GP prior $k(\vx, \vx')$. Inter-domain approximations \citep{lazaro2010sparse} simply result in a different expression for elements of $\Kuu$ and $\Kfu$, and so only require a modification to the evaluation of these matrices -- the computation of the rest of the learning objective remains unchanged. As shown in the main text, our proposed inference method is an inter-domain method, and therefore can be implemented with the same small modification. The advantage of this is that all changes to the inference can be encapsulated in the kernel.

\subsection{Exploiting convolutions}
\begin{figure}[htpb]
  \centering
  \includegraphics[width=0.5\textwidth]{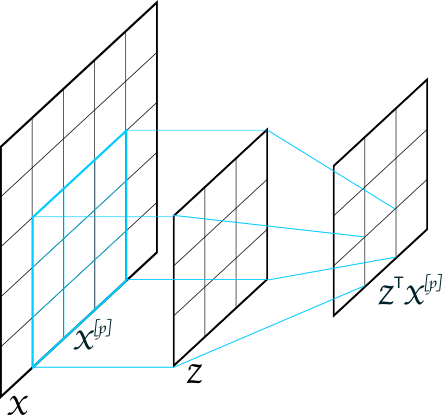}
  \caption{\label{fig:conv}Pictorial representation of convolution required in stationary kernels. A single dot product with a patch is highlighted.}
\end{figure}

A large bottleneck for the implementation is summation of kernel evaluations over numerous patches. A general implementation could simply extract the patches from the image, compute the kernel, and sum:
\begin{equation}
\left[\Kfu\right]_{nm} = k_{fu}(\vx_n, \vz_m) = \sum_p k_g(\vx_n^{[p]}, \vz_m) \,.
\end{equation}
This can be implemented as evaluating a large $PN\times M$ kernel matrix, reshaping to $P\times N\times M$, and summing over the first dimension. For general kernels, this is required. However, if the kernel is stationary, i.e.~$k_g(\vp, \vp') = k_g(\norm{\vp - \vp'})$, the first step is computing the matrix of pairwise distances between all patches and inducing points. For general inputs this still doesn't help, but in this case neighbouring inputs overlap strongly, since they're all patches from the same image. By expanding the euclidean distance as
\begin{equation}
(\xp_n - \vz_m)^2 = {\xp_n}\transpose\xp_n - 2{\xp_n}\transpose\vz_m + \vz_m\transpose\vz_m
\end{equation}
we see that the inner product of $\vz_m$ along all patches of $\vx$ is a convolution operation (figure \ref{fig:conv}). Additionally, the inner product of the patches with themselves is also a convolution of the squared image with a window of ones. This allows the euclidean distance to be computed in $\BigO\left(\log E\right)$ rather than $\BigO\left(E\right)$. An additional speed benefit comes from being able to leverage the highly optimised code for convolutions on GPUs that was developed in light of the popularity of convnets.

\section{The convolutional GP can not learn any function, but is nonparametric}
Here we show that there are sets of functions that a convolutional GP places no probability mass on. As a consequence, these functions will also never have density in the posterior, and can not be learned, regardless of the amount of observed data. This observation is particularly interesting, as the kernel still defines a nonparametric model. This places convolutional kernels in an interesting middle-ground, not seen in common examples of kernels. This constraint helps the convolutional kernel to generalise well, at the expense of possibly leaving some signal unexplained. However, as discussed in the main text, it is possible to get the best of both worlds by using a sum of a convolutional kernel and a kernel which spreads its probability mass more widely, and letting the marginal likelihood determine their relative weighting.

\subsection{Background: nonparametric models and consistency}
Nonparametric methods are usually justified by a desire to build a method that is \emph{universally consistent}, by which is meant that an optimal solution is found in the limit of infinite data. For example, in regression the unknown function may be any continuous function, and we would like our solution to be able to come arbitrarily close. Achieving this would require, at least, being able to represent any continuous function arbitrarily closely (the \emph{universal approximation property}), and for Bayesian models, a prior that places probability mass over this entire space \citep{orbanz2010bnp}.

Neural networks have been shown to be able to approximate any function arbitrary closely in the limit of having an infinite number of basis functions (i.e.~hidden units) \citep{cybenko1989universal,hornik1991nn-capabilities}. Kernel methods such as Gaussian processes, kernel regression or SVMs all implicitly use basis functions, possibly infinitely many, through their kernel. We can find a representation of the basis functions they implicitly use through Mercer's theorem \citep{mercer1909,seeger2004gpml,williams2006gaussian}, which represents a kernel in terms of its eigenvalues and orthogonal eigenfunctions:
\begin{align}
k(\vx, \vx') = \sum_{i=1}^\infty \lambda_i \phi_i(\vx) \phi_i^*(\vx') \,.
\end{align}
The span of the functions $\left\{\sqrt{\lambda_i}\phi_i\right\}_i$ determines exactly the functions that a kernel method can capture. For SVMs and kernel regression, the estimated function lies in the RKHS spanned by these bases, while the sample paths of a Gaussian process can be constructed as the infinite sum of eigenfunctions weighted by Gaussian random variables \citep[\S 5.1]{seeger2004gpml}:
\begin{align}
f(\cdot) = \sum_{i=1}^\infty w_i \phi_i(\cdot) \,, && w_i \sim \NormDist{0, \lambda_i} \,.
\end{align}

\paragraph{Degenerate kernels} We call any kernel with only a finite number of non-zero eigenvalues \emph{degenerate} (following \citep{williams2006gaussian}). Degenerate kernels can not lead to models that are universal approximators, as any function that contains a component of an eigenfunction with a zero eigenvalue, will be outside the RKHS and outside the set of functions the GP can generate. Furthermore, such a function can have an arbitrarily large deviation from functions that can be represented, simply by adding a larger component of the eigenfunction with zero eigenvalue. From this, we can easily see that an infinite number of non-zero eigenvalues is necessary for universal approximation.

We can alternatively see this by considering that a degenerate kernel with $B$ non-zero eigenvalues can be expressed as a finite basis function model. In these models, we can fully specify any function with knowledge of $B$ function values. The function value at a $B+1$th input is therefore also fully constrained. We can construct a function outside the RKHS or prior simply by adding a perturbation to the constrained function which does not pass through the $B+1$th output.

Common degenerate kernels arise from considering parametric models, like linear or polynomial models.

\paragraph{Universal kernels} \citet[def. 4]{steinwart2001} introduced the concept of a \emph{universal} kernel, which has an RKHS which is dense in the space of all continuous functions, i.e.~for every continuous function, there is a function in the RKHS with an arbitrarily small maximum deviation. Gaussian processes based on universal kernels have sample paths which are arbitrarily close to any continuous function\footnote{This follows from \citet[theorem 4]{ghosal2006}, which shows that Gaussian processes assign non-zero probability to functions that are close to functions in the RKHS of their kernel.}. The universal consistency arguments for SVM classification and kernel regression by \citet{steinwart2005} and \citet{christmann2007} rely on using universal kernels. \citet{micchelli2007universal} further characterise the properties required for universal kernels. Most common non-degenerate kernels, like the squared exponential, are also universal \citep{micchelli2007universal}.

\subsection{Convolutional kernels: nonparametric but not universal}
Here, we show that convolutional kernels fall between degenerate and universal kernels in terms of their representational capacity. We first show that we can construct a collection of inputs which fully constrains the function value at a different input, as was discussed for degenerate kernels. We then follow on to show that unlike degenerate kernels, we can still arbitrarily specify the function at an infinite number of distinct points, showing that the kernel can not implicitly be using a finite number of basis functions.

\begin{claim}
Weighted covariance kernels are not universal, and Gaussian processes based on them do not place probability on (or near) all continuous functions.
\end{claim}
\begin{proof}
Consider $W\times H$ sized images with $w\times h$ sized patches. If $w < W$ and $h < H$, we will have $P > 1$ patches in each image. There are $WH$ images with a single pixel switched on, and $wh$ distinct patches $\left\{\vz_{i}\right\}_{i=1}^{wh}$. We can organise the evaluations of $g(\cdot)$ for each of the $wh$ distinct patches in the vector $\vg \in \Reals^{wh}$, where $\left[\vg\right]_i = g(\vz_i)$. If we consider $N \leq WH$ image inputs with a single pixel switched on, we can obtain the function values $\vf \in \Reals^N$ through the linear transformation
\begin{align}
\vf = WQ\vg \,.
\end{align}
$W \in \Reals^{N\times P}$ has the patch weights as rows, and the matrix $Q \in \Reals^{P\times wh}$ contains a 1 at $Q_{ni}$ when the $n$th image contains the patch $i$, and zero elsewhere. The matrix $WQ$ has size $N\times wh$. This implies that for $N = wh + 1$, one of the function values in $\vf$ will be fully determined by the responses of the previous images. As a consequence, the kernel matrix for these inputs has to have some zero eigenvalues because the matrix
\begin{align}
\mathbb{E}\left[ \mathbf{ff}\transpose \right] &= \Kff = \mathbb{E}\left[WG \vg \vg \transpose W\transpose G\transpose\right] = WQK_GQ\transpose W\transpose
\end{align}
has rank at most $wh$, which shows that all functions with evaluations $\vf$ with a component in the null space of $\Kff$ have no density under the prior.

The construction of a singular kernel matrix $\Kff$ also implies that the kernel is not strictly positive definite, and therefore not universal \citep{sriperumbudur2011kernelrelations}.
\end{proof}

\begin{claim}
Convolutional kernels are nonparametric, in that they can not be represented as a finite basis function model.
\end{claim}
\begin{proof}
If the kernel only had a finite number of non-zero eigenvalues and the model could be expressed as a finite basis function model, all functions would admit the representation:
\begin{align}
f(\vx) = \vphi(\vx)\transpose\vw \,.
\end{align}
where $\vphi: \Reals^D \to \Reals^I$. The corresponding kernel matrix would have at most $\rank I$.

We choose an $N$ distinct images by with a distinct patch in the top left corner of the image of size $w\times h$, all other pixels being zero. Because patches overlap, we get $wh$ distinct patch responses which influence $f(\vx)$ as well as the influence from all the zero patches:
\begin{align}
f(\vx) = \sum_i g(\vx^{[i]}) w_i + w_0g({\bf 0})
\end{align}

We collect the patch responses in $G\in \Reals^{N\times wh}$, with the weights $\vw \in \Reals^{wh}$, with the image evaluations becoming $\vf = G\vw$. We obtain the covariance of $\vf$:
\begin{align}
\left[\Kff\right]_{nn'} = \Exp{g}{\sum_{i=1}^{wh}\sum_{j=1}^{wh} g\left(\vx_n^{[i]}\right) g\left(\vx_{n'}^{[j]}\right) w_iw_j} = \sum_{i=1}^{wh}\sum_{j=1}^{wh} k_g\left(\vx_n^{[i]}, \vx_{n'}^{[j]}\right) w_iw_j \,.
\end{align}
This covariance matrix can be obtained by reducing down the $Nwh\times Nwh$ covariance matrix between all patches. If we choose a universal kernel for $k_g(\cdot, \cdot)$, this matrix will always be positive definite. The reduced matrix is also positive definite since:
\begin{align}
\va\transpose\Kff\va = \sum_{nn'} a_na_{n'} \sum_{i=1}^{wh}\sum_{j=1}^{wh} \left[\Kgg\right]_{nin'j} w_iw_j = \sum_{nin'j} \left[\Kgg\right]_{nin'j} a_nw_i a_{n'}w_j > 0
\end{align}
This contradicts the model being parametric, which would allow the rank of $\Kff$ to be at most $I$.
\end{proof}

\subsection{Remarks}
The existence of non-degenerate kernels which are not universal may not come as a surprise to theoreticians, particularly due to the effort required for proving universality. For example, \citet{micchelli2007universal} place strong requirements on the form of the implicit features of the kernel, which are likely not satisfied by convolutional kernels. Despite the prescience of theory, convolutional kernels provide an interesting and practically useful example of such kernels.

\section{Variational bound for separate representation of latent GPs}
In the main text (sections 3.3 \& 3.4) we saw two examples of models with additive structure that required separate representation of their inducing outputs. The weighted convolution + RBF experiment required this due to the inducing inputs lying in separate spaces, while the multi-channel convolutional kernel required this due to separate inter-domain inducing outputs being required to find the distribution over the GP output. In both cases, we can construct the GP output of interest from components in the same space as the inducing variables:
\begin{align}
f_{sum}(\vx) &= f_{rbf}(\vx) + \sum_p w_p g(\vx^{[p]}) && f_{multi}(\vx) = \sum_{p=1}^P\sum_{c=1}^C w_{pc} g_c\left(\vx^{[pc]}\right) \\
f_{rbf}(\cdot) &\sim \GP\left(0, k_{rbf}(\cdot, \cdot)\right) && g_c(\cdot) \sim \GP\left(0, k_g(\cdot, \cdot)\right) \\
g(\cdot) &\sim \GP\left(0, k_g(\cdot, \cdot)\right) && 
\end{align}
Here we show in detail how inference is done in these models, and how no expensive operations are performed on matrices larger than $M\times M$.

\subsection{Defining the inducing variables}
We first choose our inducing variables. For the summed GP, we choose $M$ evaluations of $f_{rbf}(\cdot)$ and $g(\cdot)$ each (giving $2M$ inducing variables), while for the multi-channel GP we choose $M$ evaluations of all $C$ colour channel outputs $g_c(\cdot)$ (giving $MC$ inducing variables). We can construct the sparse approximate posterior by conditioning the prior on these variables. The form of the posterior is exactly the same as usual:
\begin{align}
f(\cdot)|\vu \sim \GP\left(\kuf\transpose\Kuu\inv\vu, k(\cdot, \cdot) - \kuf\transpose\Kuu\inv\kuf\right) \,.
\end{align}
As with usual inter-domain approximations, the task is to find the correct covariances, again by finding the covariances from equations (10) and (11).

\paragraph{Summed kernels}
For $f_{sum}(\cdot)$, the cross-covariance $k_{fu}(\vx, \vz)$ will be the regular kernel evaluation for inducing points on $f_{rbf}(\cdot)$, and the appropriate cross-covariance for the inducing patches. We order the inducing point covariances above the inducing patch covariances in the matrix $\Kfu$. Additionally, $\Kuu$ will be a block-diagonal $2M\times 2M$ matrix, with the inducing point $M\times M$ matrix for $f_{rbf}(\cdot)$ in the top left, and the inducing patch covariances in the bottom right. No cross terms between $f_{rbf}(\cdot)$ and $g(\cdot)$ appear, as they are independent in the prior.
\begin{align}
k(\vx, \vz_{img}) &= \Exp{f_{rbf}, g}{\left(f_{rbf}(\vx) + \sum_p w_p g(\vx^{[p]})\right)f_{rbf}(\vz_{img})} = k_{rbf}(\vx, \vz_{img}) \\
k(\vx, \vz_{patch}) &= \Exp{f_{rbf}, g}{\left(f_{rbf}(\vx) + \sum_p w_p g(\vx^{[p]})\right)g(\vz_{patch})} = k_{g}(\vx, \vz_{patch}) \\
k(\vz_{img}, \vz_{patch}) &= \Exp{f_{rbf}, g}{f_{rbf}(\vz_{img})g(\vz_{patch})} = 0
\end{align}

\paragraph{Multi-channel kernels}
For $f_{multi}(\cdot)$ the situation is similar, with the difference that we only have $M$ inducing inputs, but $MC$ inducing outputs. If we order the inducing variables by colour, we get an $N\times MC$ $\Kfu$ matrix (as in equation (21)), and a block-diagonal $\Kuu$, as:
\begin{align}
\Exp{\{g_c\}_{c=1}^C}{g_{c}(\vz)g_{c'}(\vz')} = k_g(\vz, \vz') \delta_{cc'} \,.
\end{align}
This process is slightly different compared to usual inducing variable approximations, and the even the case for summed kernels, as the number of inducing variables is larger than the number of inducing inputs. As a curiosity, and not necessarily a practical method of implementation, we would like to point out that we could view this process as having multi-output inducing variables. The function $g: \Reals^{wh}\to \Reals^C$ could collect all $g_c(\cdot)$s, as one $\Reals^C$ variable.

\subsection{Inference with block-diagonal $\Kuu$ matrices}
In the previous section, we saw how to find the conditional process. Here we show that the marginal likelihood bound
\begin{align}
	\textsc{ELBO} = \sum_i \Exp{q(f(\vx_i)}{\log p(y_i\given f(\bx_i))} - \textsc{KL}[q(\bu)||p(\bu)]
\end{align}
can be computed without operations on matrices larger than $M\times M$, despite using more than $M$ inducing variables, regardless of the mean-field assumptions between inducing variables.

\paragraph{Approximate posterior marginals} The bound requires computation of the marginals of the approximate posterior $q(f(\vx_i))$. This requires marginalising the conditional approximate posterior over $\vu$ (the same procedure as in \citep{hensman2015scalable,hensman2015mcmc}). This is where equations (6-8) come from. We simply substitute in the $\Kfu$ and $\Kuu$ matrices from the previous section, and simplify using the block-diagonal structure in $\Kuu$. We refer to each group of inducing variables ($C$ in total) that are correlated in a block of $\Kuu$ as $\vu_c$, and the corresponding covariance matrices $\Kucuc$ and $\vk_{\vu_c}(\vx)$. We similarly split the variational parameters $\vm$ and $\bS$ into blocks of the same size $\vm_c$ and $\bS_{cc'}$.
\begin{align}
\mu_i &= \vk_{\vu}(\vx)\transpose\Kuu\inv\vm = \sum_c \vk_{\vu_c}(\vx)\transpose\Kuuc\inv\vm_c \\
    \sigma^2_i &= k(\bx_i, \bx_i)+ \vk_{\vu}\transpose(\vx)\Kuu^{-1}(\bS-\Kuu)\Kuu^{-1}\vk_\vu(\vx) \nonumber \\
&= k(\vx_i,\vx_i) + \sum_{cc'}\vk_{\vu_c}(\vx)\transpose \Kucuc\inv \bS_{cc'}\Kucuc\inv \vk_{\vu_c}(\vx) +  \sum_{c}\vk_{\vu_c}(\vx)\transpose\Kuuc\inv\vk_{\vu_c}(\vx)
\end{align}
In both cases outlined above, $\Kucuc$ is $M\times M$. If a mean-field approximation is chosen $\bS_{cc'} = 0$ when $c \neq c'$. This does not impact the number or size of any inverses, only requiring less parameters and avoiding a summation over $c'$.

\paragraph{KL divergence} The second term in the bound requires the KL divergence between the prior and posterior distribution over the inducing variables, which we can again simplify using knowledge of the block-diagonal structure.
\begin{align}
&\KL{q(\vu)}{p(\vu)} = \frac{1}{2}\left(\Tr\left(\Kuu\inv\bS\right) + \vmu\transpose\Kuu\inv\vmu - MC + \log \frac{\detbar{\Kuu}}{\detbar{\bS}}\right) \\
& \qquad = \frac{1}{2}\left(\sum_c\Tr\left(\Kuuc\inv\bS_{cc}\right) + \sum_c\vmu_c\transpose\Kuuc\inv\vmu_c - MC + \sum_c \log \detbar{\Kuuc} - \log \detbar{\bS}\right)
\end{align}
Now, the determinant of $\bS$, which may be of size $MC\times MC$ remains. Luckily, we are free to choose the parameterisation of this matrix. We parameterise this matrix as $\bS = LL\transpose$, which makes $\log\detbar{\bS} = 2\sum\log\diag L$.


\subsection{Summary}
Here we showed that when the prior covariance of the inducing outputs is block-diagonal, the inference requires only requires expensive matrix operations on each of the blocks separately, regardless of the posterior correlations taken into account. This allows efficient inference for the summed and multi-channel convolutional kernels considered here.

\section{Inter-domain inducing variables for general invariances}
We finally briefly show that the inter-domain trick used for convolutional kernels can also be applied to kernels that give rise to Gaussian processes with arbitrary invariances. Invariant kernels have been discussed before, notably by \citet{kondor2008thesis} and \citet{ginsbourger2012,ginsbourger2013,ginsbourger2016}. \citet{duvenaud2014thesis,duvenaud2014kernelcookbook} also provides an accessible discussion. Here, we review the connection between kernels resulting in invariant functions and a summation structure which allows our inter-domain trick to be applied.

\subsection{Specifying invariances in kernels}
An invariance can be formalised by placing equality constraints on $f(\cdot)$ under transformations of the input. Consider a collection of transformations from the input space to itself $g_i:\mathcal{X} \to \mathcal{X}$. Making $f(\cdot)$ invariant to these transformations specifies that
\begin{align}
f(\vx) = f(g_i(\vx))  && \forall\vx \in \mathcal{X}  && \forall i \,.
\end{align}
\citet{kondor2008thesis} and \citet{ginsbourger2012} discuss that this requirement is equivalent to invariance under every composition of transformations as well. For example, if $g_1(\cdot)$ and $g_2(\cdot)$ are translations upwards and to the right respectively, we must also have invariance to a translation up and to the right $f(\vx) = f(g_1(g_2(\vx)))$. The set of compositions of all transformations forms a group $G$. \citet[theorem 3.1]{ginsbourger2012} show that in order for samples $f(\cdot)$ to be invariant to all compositions of transformations, the kernel must be \emph{argumentwise invariant}:
\begin{align}
k(\vx, \vx') = k(g(\vx), g'(\vx')) && \forall\vx,\vx'\in\mathcal{X} && g,g'\in G \,.
\end{align}
The elements of the group $g\in G$ are all compositions of the transformations $g_i$, defined above.

\subsection{Constructing invariant kernels}
The requirement stated above does not directly help with constructing invariant models. Three main methods have been proposed, which are neatly discussed by \citet[\S 2.7]{duvenaud2014thesis}. For our purposes, we are mainly interested in the ``summation over orbit'' method, as this gives a structure almost identical to the convolutional kernel.

\citet{kondor2008thesis} and \citet{ginsbourger2012} show that an argumentwise invariant kernel can be constructed by summing some base kernel over the \emph{orbits} of $\vx$ and $\vx'$. The orbit of a point $\vx$ with respect to a group $G$ is defined as the set of all points obtained from applying each element of $G$ to $\vx$: $\mathcal{O}_G(\vx) = \left\{g(\vx)\mid g\in G\right\}$. The resulting kernel becomes:
\begin{align}
k_{invariant}(\vx, \vx') = \sum_{\tilde{\vx}\in\mathcal{O}_G(\vx)} \sum_{\tilde{\vx}'\in\mathcal{O}_G(\vx')} k_{base}(\tilde{\vx}, \tilde{\vx}') \,.
\end{align}
The relation between invariances and the addition structure is further investigated by \citet{ginsbourger2013add}.

Kernels constructed in this way have the same computational issues as convolutional kernels: evaluating the kernel for a single pair of points requires $P^2$ base kernel evaluations, where $P$ is the size of the orbit. For example, we could make a fully translation invariant kernel by considering translations by 1 pixel upwards, downwards and to the left and right, while clipping and zero-padding edges. For images of size $W\times H$ the orbit would consist of all $W-1\times H-1$ translated images. For MNIST this would give $P^2 = (27\times 27)^2 = 5.3\cdot 10^5$, which is again impractical.

\subsection{Inter-domain inducing variables for invariant kernels}
The invariant kernel above can also be obtained by considering a model that sums a base function $f_{base}(\vx) \sim\GP\left(0, k_{base}(\cdot, \cdot)\right)$ over the orbit of $\vx$:
\begin{align}
f(\vx) = \sum_{\tilde{\vx}\in\mathcal{O}_G(\vx)} f_{base}(\tilde{\vx}) \,.
\end{align}
In this construction, the base function $f_{base}(\cdot)$ takes the place of the patch response function $g(\cdot)$ from the convolutional kernel, allowing us to use the same inter-domain trick. Instead of using normal inducing inputs, we place the inducing inputs in $f_{base}(\cdot)$ instead. We then obtain the covariances:
\begin{align}
k(\vx, \vz) &= \sum_{\tilde{\vx}\in\mathcal{O}_G(\vx)} k_{base}(\tilde{\vx}, \vz) \,, \\
k(\vz, \vz') &= k_{base}(\vz, \vz')\,.
\end{align}

Just like with the convolutional kernel, this reduces the cost of evaluating the required kernels significantly.

\subsection{Summary}
The structure of kernels resulting in GPs that are invariant to specified transformations is almost identical to that of convolutional kernels, allowing the same inter-domain trick to be used to speed up inference. We present the derivation here, but leave empirical demonstration and evaluation to future work.


\bibliographystyle{unsrtnat}
\setlength\bibsep{0pt}
{\small\bibliography{references}}